# Learning Syntactic Rules and Tags with Genetic Algorithms for Information Retrieval and Filtering: An Empirical Basis for Grammatical Rules*

Robert M. Losee
Manning Hall, CB #3360
U. of North Carolina
Chapel Hill, NC, 27599-3360
USA

losee@ils.unc.edu

**Abstract**

The grammars of natural languages may be learned by using genetic algorithms that reproduce and mutate grammatical rules and part-of-speech tags, improving the quality of later generations of grammatical components. Syntactic rules are randomly generated and then evolve; those rules resulting in improved parsing and occasionally improved retrieval and filtering performance are allowed to further propagate. The LUST system learns the characteristics of the language or sublanguage used in document abstracts by learning from the document rankings obtained from the parsed abstracts. Unlike the application of traditional linguistic rules to retrieval and filtering applications, LUST develops grammatical structures and tags without the prior imposition of some common grammatical assumptions (e.g., part-of-speech assumptions), producing grammars that are empirically based and are optimized for this particular application.

---

*The author wishes to thank Stephanie Haas for discussions during the course of this research.

# 1 Introduction

A language contains a set of terms and rules capable of manipulating the terms, producing the "grammatical" statements permitted by the language. Using the LUST (Linguistics Using Sexual Techniques) genetic algorithm system, we have developed and evaluated grammars based, not on their degree of similarity to grammars intellectually developed by human grammarians, but, instead, on how well a system performs a task using the genetically developed grammar. The progressive improvement of the grammar produced by LUST will gradually increase the performance of the retrieval task on which LUST was trained. The grammar used here to parse natural language text is represented as a "gene" that evolves through a process similar to that found in natural selection, where those creatures in an environment that have an advantage over other creatures are more likely to survive and proliferate. LUST allows for rules to be generated, propagate, or "die off," depending on how well the retrieval or filtering task is performed that uses LUST's grammatical rules and part-of-speech assignments. In a document retrieval or filtering system, applying grammatical tags to the list of terms representing a document provides additional information about the semantic content and structure of the document that is not present in the untagged document. One parse may fail to distinguish between two different uses for the same term, resulting in conventional retrieval performance, while an improved parse may note the distinction and produce better retrieval than would be produced by the first parse.

Information retrieval and filtering systems can retrieve or predict the usefulness of document information given bibliographic descriptions of the document. These retrieval and filtering systems may be studied using traditional retrieval and filtering performance measures such as precision and recall (Salton & McGill, 1983; Van Rijsbergen, 1979) or measures such as average search length (Losee, 1991). Each retrieval performance measure can serve as a fitness function for a genetic algorithm system, such as LUST, providing a measure by which grammatical rules developed by LUST may be evaluated. Those rules whose application results in improved disambiguation are expected to produce improved retrieval and filtering performance, or, in the worst case, we expect that such knowledge will not hurt retrieval performance.

Human grammarians usually assume that individual terms used in natural language may be grouped into categories, which we refer to as *parts-of-speech,* and then labeled with *grammatical tags,* and that linguistic rules can be used to manipulate any term having certain characteristics. Commonly accepted parts-of-speech include such categories as *verb*, *object*, and *noun*, among many others. The basis for defining and using these grammatical components has always been a loose one, based on how well the grammatical categories seem to "fit" grammarians' intuitions about the fundamental structure of a language.

With the formalization of syntactic rules by Zellig Harris (Newmeyer, 1986) and Noam Chomsky (1965), various sets of syntactic rules and criteria for such rules have been proposed and can serve as the basis for the grammars described below. One may also examine the parsing or understanding of phrases and small parts of sentences, in lieu of parsing entire sentences. Some approaches describe the statistical relationships existing between terms using hidden Markov models (Charniak, 1993). Other work emphasizes the examination of phrases or term clumps bounded by term windows (Haas & Losee, 1994; Losee, 1994). The research described below begins by viewing sentences in terms of their grammatical components rather than by taking a more statistical approach. Future research will attempt to study more limited structures that the author believes may be learned and modeled more precisely than can whole sentences.

Genetic algorithms can be used to learn the characteristics of a wide variety of phenomena, both inside and outside linguistic and document retrieval domains. In addition to being applied in a variety of biological and industrial environments, they have been used to model and study the historical changes in a language (Clark & Roberts, 1993). The nature of linguistic phenomena may be learned through the application of other techniques, such as neural networks capable of learning to associate events with other events. Genetic algorithms were chosen for this work because at each stage in the evolutionary process, a full grammar and set of part-of-speech tags is provided by the gene, making easier the qualitative and quantitative evaluation of the derived language.

In this work, we empirically study how genetic algorithms may be used to assign parts-of-speech tags to a set of terms. These part-of-speech assignments are allowed to evolve, with those part-of-speech assignments leading to gains in the value of a fitness measure (e.g. improved retrieval results) contributing to a greater chance of the part-of-speech assignment surviving. Those part-of-speech assignments resulting in poor retrieval are less likely to survive and are instead replaced by other part-of-speech assignments.

## 2   Syntactic Genes

A grammar describing a language may be understood as consisting of a set of grammatical rules and a set of part-of-speech tags for terms. These syntactic rules and the relationships between grammatical tags and terms may be stored as alleles, individual elements within genes, for analysis with a genetic algorithm. A rule, such as $A \to BC$, may represent grammatical component $A$ being composed of grammatical components $B$ and $C$, in that order. (Within LUST, there are provisions for using the IDLP grammar described in (Gazdar, Klein, Pullum, & Sag, 1985), which uses some unordered rules unlike those used in more traditional studies of syntax and is attracting increasing interest in the linguistics community (Briscoe & Carroll, 1993; Chitrao & Grisham, 1990)).

For the purposes of the LUST system, each gene contains a set of syntactic rules (a constant number of rules for each possible part-of-speech on the left hand side of a syntactic rule). Each non-terminal symbol on the left hand side of a rule could have $n$ different rules describing its direct composition, where $n = 5$ is the default value for the experiments described below. Each term has two (not necessarily different) part-of-speech assignments. Thus, the term *run* might evolve to being tagged as a verb or a noun or both. Note that for our purposes, part-of-speech labels are arbitrary and have no "meaning" to the system. The system does not label a term as being a *noun,* for example, or a *verb*; instead, a *label number* is attached to several terms. These may be understood by a human to be members of a certain category of term. For example, all terms of category 5 might be understood by a grammarian to be verbs, but LUST does not need or use this information.

Syntactic production rules may vary in the number of terms on the right hand side of the rule, or each production rule may have a fixed number of terminal and non-terminal symbols, e.g. 2. LUST features may be produced by a truncated Poisson process, generating an average of 1.8 terms per rule (an arbitrarily chosen number) and never producing 0 terms. There frequently will be 1 or 2 terms per rule, and seldom over 5 terms in a rule. A similar distribution has been recently proposed by Lankhorst (1994b).

## 3 Parsing and the Retrieval Process

The parsing and retrieval process in the LUST system begins with the parsing of each sentence with a chart parser (Charniak, 1993). Given a set of grammatical rules and part-of-speech tags, the chart parser produces the set of parses that can produce the sentence. Disambiguation is accomplished by selecting the parse produced by applying the fewest number of rules. The chart parser produces for each term the accompanying grammatical tags, each term having attached to it all those grammatical markers found when moving up a conceptual parse tree for $n$ levels. Thus if $n = 1$, the term *dog* in *dog bites man* might be stored as the unit *dog—noun* while if $n = 2$ (the LUST default) it might be stored as *dog—noun—subject*. Each *term complex* is treated as a unit, and *dog—noun—subject* is treated as a different term complex than *dog—noun—object*. In most cases below, it is obvious whether we are describing an individual term or a term complex and thus refer to both as a term; when confusion might arise, we will use either the expressions "term" or "term complex." Note that $n = 0$ provides essentially untagged terms, such as is found in document retrieval or filtering systems having no linguistic knowledge.

LUST uses document abstracts for parsing and retrieval. In some experiments described below, only the first sentence of each abstract is used, while in others, all sentences are used. Document titles are not currently used. Often, multiple parses are obtained for each grammatical component and disambiguation is used to select one of the parses as the parse to be used for the remainder of the processing.

Documents are weighted using a traditional probabilistic weighting formula (Robertson & Sparck Jones, 1976; Bookstein, 1983; Losee, 1988). Each document $j$ is assigned a value based on the sum of term or term complex weights, based on probabilities of terms occurring in a given class of documents:

$$RSV_j = \sum_{i=1}^{n} d_i \log \frac{p_i/(1-p_i)}{q_i/(1-q_1)},$$

where $p_i$ is the probability that a binary feature $i$ is present in relevant documents, and $q_i$ is the probability that binary feature $i$ is present in non-relevant documents. Documents are ranked for retrieval or examination in order of their retrieval status value (RSV). The average search length (ASL), the average number of documents retrieved when retrieving a document in the average position of a relevant document, may then be computed from this ranking and from knowledge the system has about the relevance values for each document (given the query).

When a term complex contains more than the single term, that is, it contains a grammatical tag, the same term may occur in different complexes. Probabilities of term complexes may be difficult to estimate accurately, since, for example, there may be two types of *dog* occurrences, one as subject, and one as object, both occurring with a lower frequency than the untagged term *dog*. In a realistic retrieval or filtering system, these probabilities must be estimated, and having fewer occurrences of a term complex (as compared to what would be obtained in conventional systems without tagging) will usually result in less accurate estimates because of fewer data points being included when making each estimate. It remains to be seen whether making lower quality estimates of probabilities of more accurately tagged text features will result in improved or possibly even diminished performance (Losee, Bookstein, & Yu, 1986).

## 4 Evaluative Criteria

When modeling evolutionary processes using genetic algorithms, it is necessary to use a fitness function to evaluate the performance or "fitness" of an individual gene. Those genes that are most fit are most likely to survive, with less fit genes dying off, being replaced by the fitter genes. Several possible functions may be used in determining the fitness and efficacy of a grammar. One function that the author has used previously (Losee, 1994) is the average search length (ASL). Evaluating the performance of a filtering or retrieval process with the ASL satisfies many of the needs of retrieval researchers, in that it provides a single number measure of performance, the ASL is capable of being predicted analytically (Losee, 1995), and the ASL is easily understood by a system's end users. This is unlike most of the other retrieval and filtering measures based on precision and recall that are used to evaluate retrieval systems. Other measures may be useful in determining the fitness of a grammar. One measure is the number of rules applied during the parsing process. Another fitness function is the average (over a set of sentences) of the largest number of terms in a parse for each sentence, the average maximum parse length (AMPL).

Because properly assigning the part-of-speech to a term is not expected to improve greatly retrieval and filtering performance, measures combining the ASL and a second measure that does not depend on retrieval performance may be useful. In the work below, a useful fitness measure was obtained by weighting the ASL value and the AMPL so that one hundredth of the AMPL was added to the negation of the ASL. The greater the value for this combined function, the fitter the gene that produced it. It is common for different genes to produce the same ASL value, and adding the AMPL allows for a finer level of discrimination between genes. The modification acts to break ties that exist when using ASL alone as the fitness function. When two genes with identical ASLs are compared, the gene with the larger AMPL is selected. This weighting has the side effect of selecting a new gene if the new gene has an inferior ASL but a much superior maximum parse length. This computational method is useful when the ASL is not a very good measure of the quality of parsing performance due to the limited improvements obtained with disambiguation.

## 5 The LUST System

The LUST system has been developed as a rapid prototyping system that can be easily modified as the experimenter gains insights into the application of genetic algorithm techniques to linguistic analysis. All code was written using Unix Bourne shell scripts and gawk, a version of awk. Execution time for programs was slower than would have been obtained with code in a lower level language such as C or C++. However, Unix shell scripts are particularly efficient at handling natural language text such as that studied here, allowing the programs to be easily coded and modified for different tests.

A database of 108 abstracts was developed containing 988 sentences. These abstracts are in five groups, each being an extract from one of five larger databases on psychology developed by Stephanie Haas for sublanguage analysis (Losee and Haas, 1994). Each of the five groups was originally retrieved with a single query, thus all documents in a group are similar in a topical sense. Documents abstracts were manually placed into a regular linguistic form, with parenthetical comments removed in some cases, abbreviations expanded, spellings made consistent, etc.

The approximately twenty documents in each of the five groups, for a total of 108, were treated as "relevant" to an information retrieval query. Thus, documents in group 1 were considered relevant (and documents in the other groups non-relevant) to a particular query, and the documents' rankings were then mea-sured. This was done iteratively for documents in each of the other four groups, and the performance figures for retrieving the relevant documents in each of the five groups was averaged. Performance was measured by the ASL.

Terms used in the retrieval process are labeled by LUST as to their part-of-speech. Labels are arbitrary, indicating for example that two terms such as *run* and *walk* have a common part-of-speech, rather than indicating that a term has a part-of-speech that might be found in a grammar text (e.g. verb). To simplify the grammar, only ten parts-of-speech were used (and twenty non-terminal symbols were used).

Mutations take place in LUST by producing new rules, combining fragments from the right hand side of "fit" rules with the same left hand side as the rule it replaces, as well as occasionally producing fragments with random contents. Those "fit" rules that contribute fragments toward producing the new rule are referred to here as the "parents," with the new gene being referred to as an "offspring" or "child." Rules that are most useful when the parents' grammars serve as the basis for the parse may be included with greater frequency in whole or in part in the offspring, while rules that are seldom used may be more likely to be replaced with a new, random rule (with the same left hand side as the rule it replaces). A similar procedure may be used for the rules governing terms and their dominating non-terminal symbols (grammatical tags). This would have the effect of weighting the offspring by those parts of a parent that have been most useful to the parent (Clark & Roberts, 1993). To simplify the interpretation of genes in LUST, the results described below were derived from a system with a constant probability assigned to the possibility of each rule changing

There are always three genes in use by the LUST system, numbered 1, 2, and 3. Genes 1 and 2 are always parents, genes found previously to be the most fit, while gene 3 is the offspring produced by the parents mating. The fitness of gene 3 is determined by examining the sentences being parsed, the document rankings, and the resulting performance). After assigning a fitness value to Gene 3, the three genes are then sorted in order of the value of each of their fitness functions to get an ordered set of genes, with the fittest two becoming the parents of the next generation.

Unlike some evolutionary systems, LUST only produces a single offspring. Each gene contains a set of alleles, or rules, and when offspring are produced, the offspring contain some alleles from one parent and some from another. For example, a human child may inherit the shape of its ears from one parent and hair color from another parent. In LUST, an offspring has some of the characteristics of each parent, with some characteristics being randomly generated. The parsing function takes up the vast majority of the LUST system's time, and evaluating the fitness function (parsing) is to be minimized because parsers are generally not very fast. Rather than two parents having two offspring, each of which would need to be fully evaluated, we produce only a single offspring, which has the capability to be incorporated into the process of producing the next offspring if it is a fitter gene. This introduces superior genes into the parsing process more quickly than if two genes were produced each generation. This increases the rate of learning, although valuable rules from one of the parents can be lost if there are other rules in the children that are "better" than the lost rule.

The initial grammar is made using random processes that produce rules consistent with the procedures above and the parameters of each particular experiment. Copies of this initial grammatical gene become both parents and the initial offspring when the system begins running. The offspring is then evaluated and the mutation/reproduction – parsing – retrieval cycle begins.

# 6 Results

Results from simulations using the LUST system suggest that the system can learn information about linguistic structures, as evidenced by the consistent increase in retrieval performance as syntactic genes evolve. Each simulation runs for a particular number of generations, with the probability that a syntactic rule will be changed during a generation having a default value of .20 and the probability that a term will be assigned a new part-of-speech tag having a default value of .20. Two part-of-speech tags are assigned to each term and 5 rules govern the possible productions or transformations that can take place for each non-terminal symbol. The default number of terms on the right hand side of a syntactic rule was 2.

Each figure shows the smoothed values for a data set from the first generation up to the last useful value obtained in the simulation. These graphs compare the average maximum parse length (AMPL) over the set of databases and the number of generations that have elapsed, up to the last generation for which we know that all data is available. The last performance value that was obtained in each simulation might (or might not) have been the value for the next generation if the simulation had been allowed to continue. For this reason, the average generation for a particular performance level could not be determined for the final generation and these values were not used in determining the performance data graphed here. This data was truncated, having the effect of making some plots appear shorter than others, although results are from the same experiment.

All the experimental results may best be viewed against a baseline AMPL obtained from randomly generated genes of 3.9 to 4, with the variance being due to variations in system options. Similarly, the ASL obtained when no parsing is used is 49.914. This can be interpreted as saying that the user would need to examine an average of 49.914 documents when getting to the average relevant document in this test database.

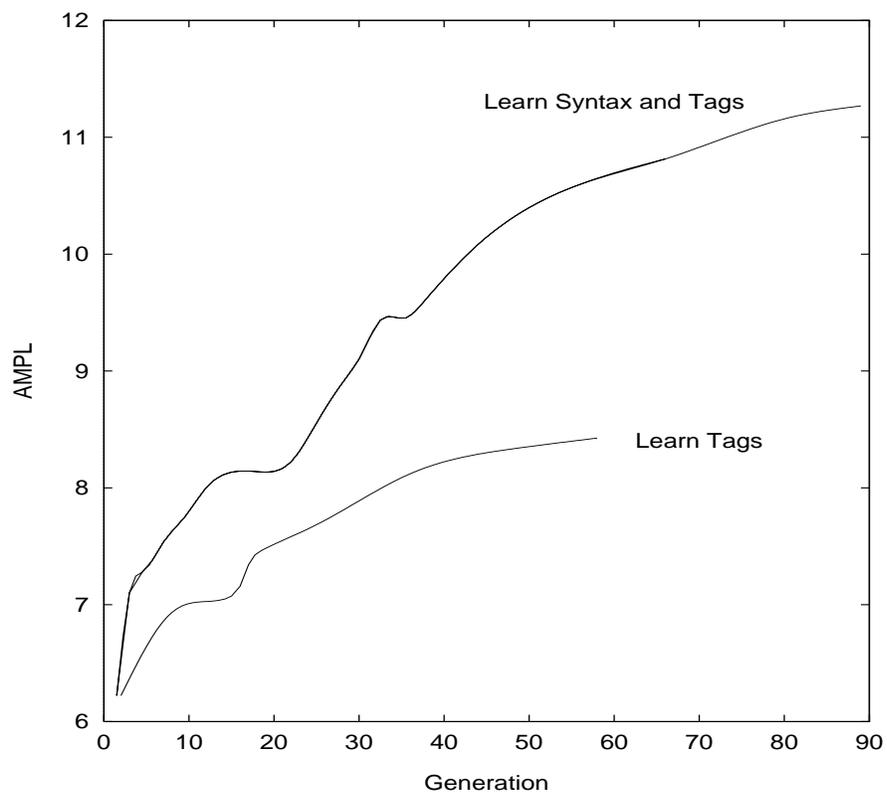

Figure 1: Learning of syntactic rules alone, part-of-speech tags alone, and both types of rules. The data for learning syntactic rules alone runs concurrently for much of this data set with learning syntactic rules and tags together.

## 6.1 Learning Syntactic Rules and Part-of-Speech Tags

Figure 1 shows the results from one set of simulations where only syntactic rules were learned and allowed to mutate, that is, where the tag values were kept constant. Another simulation allowed only the part-of-speech component to be learned, forcing the syntactic production rules to remain constant. A third experiment shows what happens when both syntactic and tagging rules are learned. The figure shows that learning either the tagging rules or the syntactic rules separately does result in improved parsing. Learning both tagging rules and syntactic rules together appears to add little to learning the syntactic rules alone for this case, but since learning tagging rules alone does improve performance, it was decided to learn both syntactic rules and tags for the experiments below.

## 6.2 Parts of Speech

These experiments were intended to learn grammatical components "from scratch," that is, they use virtually no prior knowledge about the parts-of-speech of individual terms and the syntactic rules of the language. One area of interest is how much assistance is provided to the parsing process when some limited prior information is available. This was done by labeling all the following terms on each line below as being of the same part-of-speech when the first gene is produced:

1. above below on in near through with
2. or and
3. I you he she it we they
4. a an the
5. *all numeric data*

The system does not tell the parser that *below* or *near* are prepositions; instead these terms are only initially assigned the same part-of-speech. The parts-of-speech for each of these term groups may change during the mutation process. These initial similar part-of-speech assignments merely coax the evolutionary system in a direction given some assumptions that we feel won't be too controversial.

An analysis of the labeling of these terms after a run involving fifty generations found that at least half of the rules for term's part-of-speech label in each of the categories above were still grouped together with the original label. That these terms remained with their original label is probably not significant; what is important is that the similar terms remain clustered together. Given that 20% of the part-of-speech labels mutated each generation, one would expect far fewer of the terms to remain clumped together into similar parts-of-speech after the fifty generations.

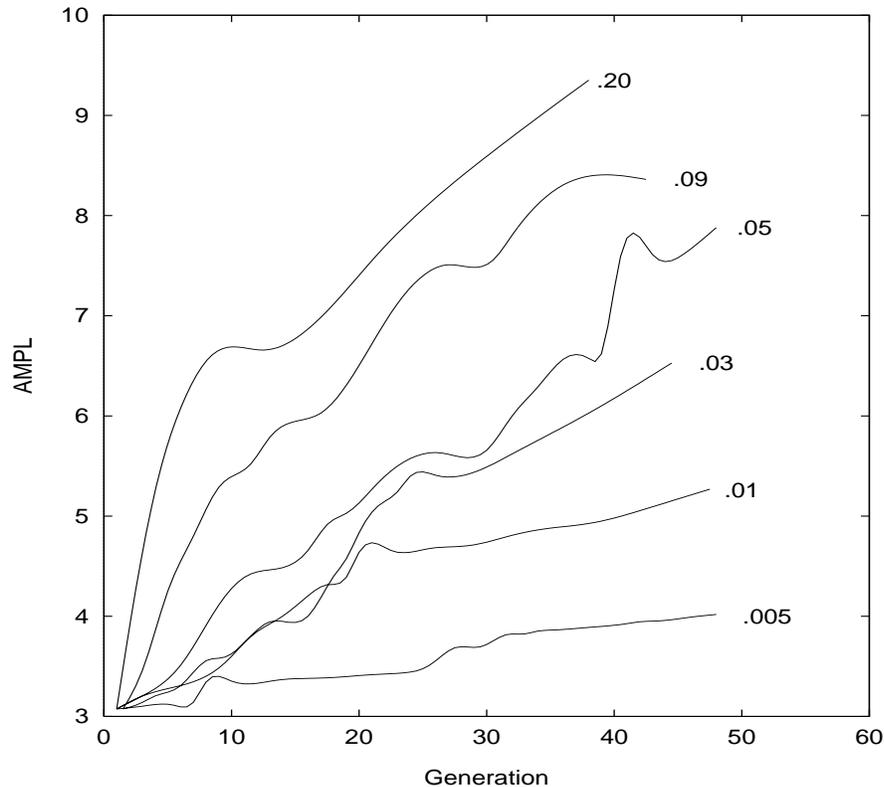

Figure 2: The rate of performance improvement increases as the mutation rate rises.

### 6.3 Mutation Rate

A separate set of simulations examined what happens when the mutation probability for either a syntactic rule or a tagging rule is allowed to vary. Only the first sentence from each abstract is used for these tests to speed up the system, producing results in the time available with the hardware available.

Figure 2 shows how parsing performance varies as the probabilities that a tag or syntactic rule can mutate is allowed to vary from .005 to .20. The performance increased more rapidly as the probability of change was allowed to increase, suggesting that as the mutation rate increases, the filtering results produced by using the genes more rapidly increase and approach their final values. Intermediate values, not shown in the figure, support this general trend, although there are situations where the performance for one probability of mutation breaks from this trend for several generations. The general trend was something of a surprise to the author, as anecdotal evidence about other genetic algorithm system designers suggested that much smaller probabilities worked better in other tested evolutionary environments.

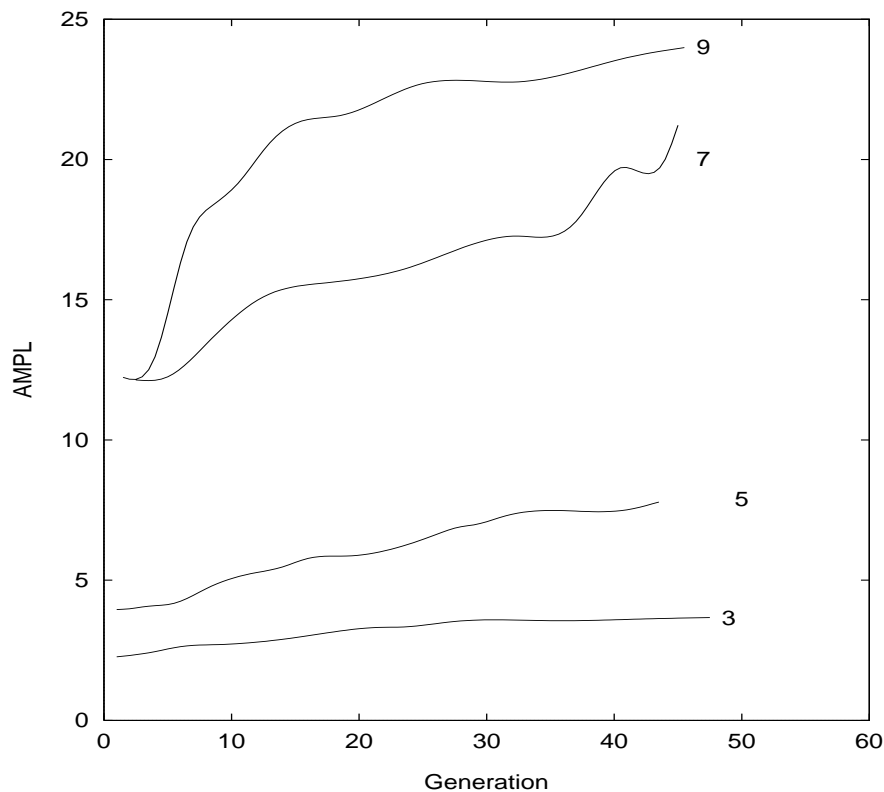

Figure 3: Performance improves as the number of rules per left hand side increases.

### 6.4 Syntactic Rules

Figure 3 shows how parsing performance improves as the number of syntactic rules for which a particular part-of-speech is in the left hand side of the production rule is allowed to vary from 3 to 9. Fifty generations of mutations were used to produce these results. The probability that a given rule or term tag was changed during a mutation was set to .03. The figure shows that as the number of right hand sides per left hand side increases, that is, the number of productions or transformations per non-terminal symbol increases, the performance improves. This is due in part to the increased number of possible rules with which to parse a sentence.

The experiments described above all use two non-terminal symbols on the right hand side of each syntactic rule, except for those that involve a single non-terminal symbol on the left and a term on the right (tagging rules). While these fixed rule formats allow for some degree of control in experiments, most grammatical systems proposed by linguists use varying numbers of non-terminal symbols on the right hand sides of syntactic rules.

If the number of terms is allowed to vary, we may model the number of non-terminal symbols as described by the Poisson distribution. This was an arbitrary distribution chosen by the author. Recent research by Lankhorst (1994b) has used a similar model for the distribution of non-terminal symbols. Our work has produced a set of non-terminal symbols where this number is Poisson distributed with the average number of terms generated being 1.8. The Poisson distribution produces some instances where zero terms will be generated. To avoid this problem, a new number of terms is generated if a 0 is produced by the Poisson random number generator. The distribution is a truncated Poisson distribution. Further research will involve testing this distribution and determining appropriate parameter values.

## 6.5 Fitness Measures

The relative performance of LUST was examined using different fitness measures to evaluate the quality of a gene. It was erroneously expected that the retrieval performance would serve as an adequate fitness function. Knowing the parts-of-speech and using the resulting disambiguation improved retrieval performance a very small amount. For this reason, measures besides raw retrieval performance (ASL) were used.

A second fitness measure is the average number of terms in the largest phrase parsed from each sentence. the *average maximum parse length* (AMPL). A third measure is a weighted combination of the first two measures, with the weighting effectively choosing the ASL alone in most cases. In the case of a tie in ASLs, the gene that produces the larger parse receives the higher weight.

Experiments here began with the same gene, and each gene then evolved through 100 generations. Using the ASL alone resulted in the gene progression from an ASL of 49.92 to an ASL of 49.90, a relatively small improvement in retrieval performance. At the same time, the AMPL increased from 4.656 to 6.034. The ASL level of 49.90 was reached by generation 4, and no better performance is obtained with additional mutated syntactic rules.

When the AMPL alone was used as the fitness function, the AMPL increased from 4.656 to 10.092, with the latter being reached at generation 61. The ASL went from 49.92 to 49.9, with the latter being obtained from generations 6 to 14 and from generations 27 to 100.

When the ASL was combined with the AMPL so that ties in the ASL were broken by the AMPL, the AMPL increased from 4.656 to 9.912 with the ASL moving from 49.92 down to 49.9 with a considerable amount of fluctuation.

The number of successful parses (NSP), the number of rules that were applied when parsing a set of sentences, was examined as a fitness function. It was noticed that while using the AMPL as a fitness function, a new value was often obtained for the AMPL at the same time that one of the highest values was obtained for the NSP. This NSP value associated with the best AMPL had usually only been exceeded once or twice before and it is obvious that there was a strong relationship between AMPL and NSP. If this is the case, and an NSP this high had occurred previously, it was reasoned that perhaps using NSP as the fitness function would result in the system learning from these earlier high NSP genes rather than from early genes with lower NSPs that had the (then highest) AMPL.

In a sense, the NSP can be understood as having a finer grain than the AMPL, with the NSP assigning a greater range of values to genes, allowing fine degrees of difference to be rewarded or discouraged, through evolutionary techniques, that would otherwise not be detected with a coarser grained measure like AMPL. If the AMPL and NSP measures are strongly (but not perfectly) related, it is likely that using the NSP will result in a more rapid development of fitter genes. Using the AMPL as a fitness function results in the two parents being essentially randomly selected from the moderately sized pool of genes with this particular AMPL value. Using the NSP, however, results in more rapid changes in parents due to the smaller (and better) pool of parents from which offspring are produced. Parents at any point can be expected to be slightly better than would be the case with AMPL, resulting in more fit offspring and, in general, more rapid learning.

Using the same initial parent genes, a sample run produced seven different AMPL values during the first fifty generations when AMPL was used as the fitness function. However, when using NSP as the fitness function, fifteen different AMPL were observed. In addition, the AMPL of 9.35223 was obtained after the first hundred generations when AMPL was used as the fitness function; a better AMPL of 9.50202 was achieved during the same period when NSP was used as the fitness function. This provides some preliminary evidence that NSP may be a better fitness function for LUST.

## 7 Derived Syntactic Rules

The results presented in the preceding section suggest that the LUST system produces linguistic rules that are superior to the random rules with which each run begins. For example, the AMPL may be easily doubled over the first one hundred generations, evidence that superior rules are being produced through the application of evolutionary processes. However, the rules that are produced after one or two hundred generations as a result of mutation are nowhere near as good as those that would be produced by a human (or as might be produced by a larger system with more data and with access to greater computational power).

Changes in the parsing ability of the system can be observed after evolution occurs, although the improvements are difficult to characterize in a systematic way because they are so irregular. We had originally expected that very widely applicable rules would be learned first, such as that a sentence may be constructed of a noun phrase and a verb phrase (e.g. NP=*The girl* VP=*climbed the tree*), or that a prepositional phrase may be composed of a preposition followed by a noun phrase (e.g. *on the table*).

Instead, the rules that developed were far less general and far more difficult to interpret. Figure 4 shows a parse tree for part of a sentence after one hundred generations of mutation, with an AMPL of about 10. An examination of this parse tree shows relatively little that corresponds to the grammatical rules that a rational human would produce! After another one hundred generations, the best gene to that point produced the parse shown in Figure 5 for the same part of a sentence. While the syntactic rules are obviously very different, as is indicated by the different numbers representing arbitrary grammatical components, the structures have an obvious similarity.

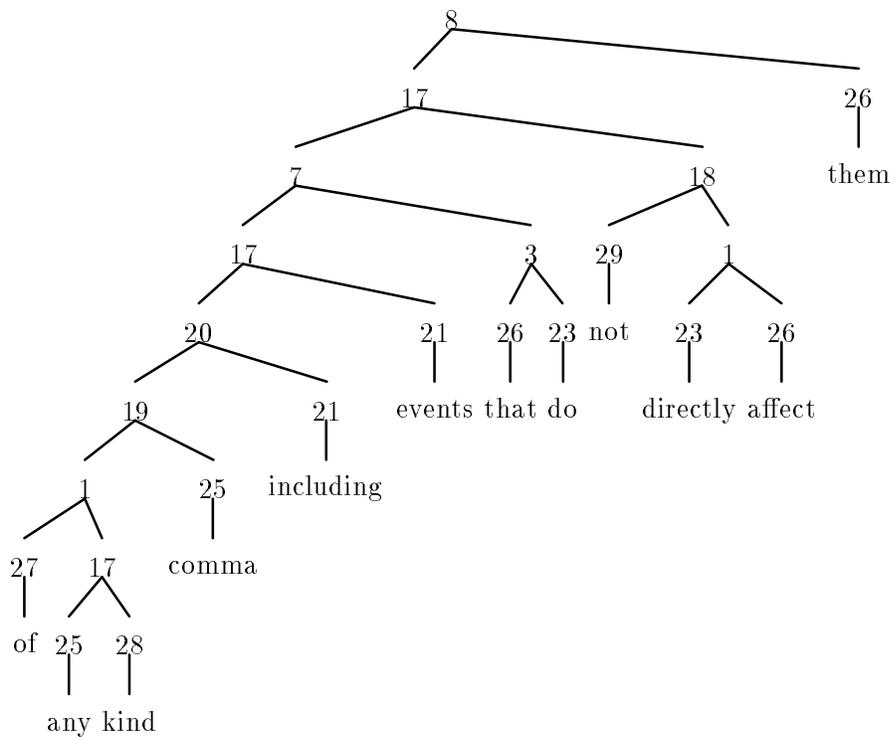

Figure 4: A parse tree for part of a sentence. This was produced after one hundred generations.

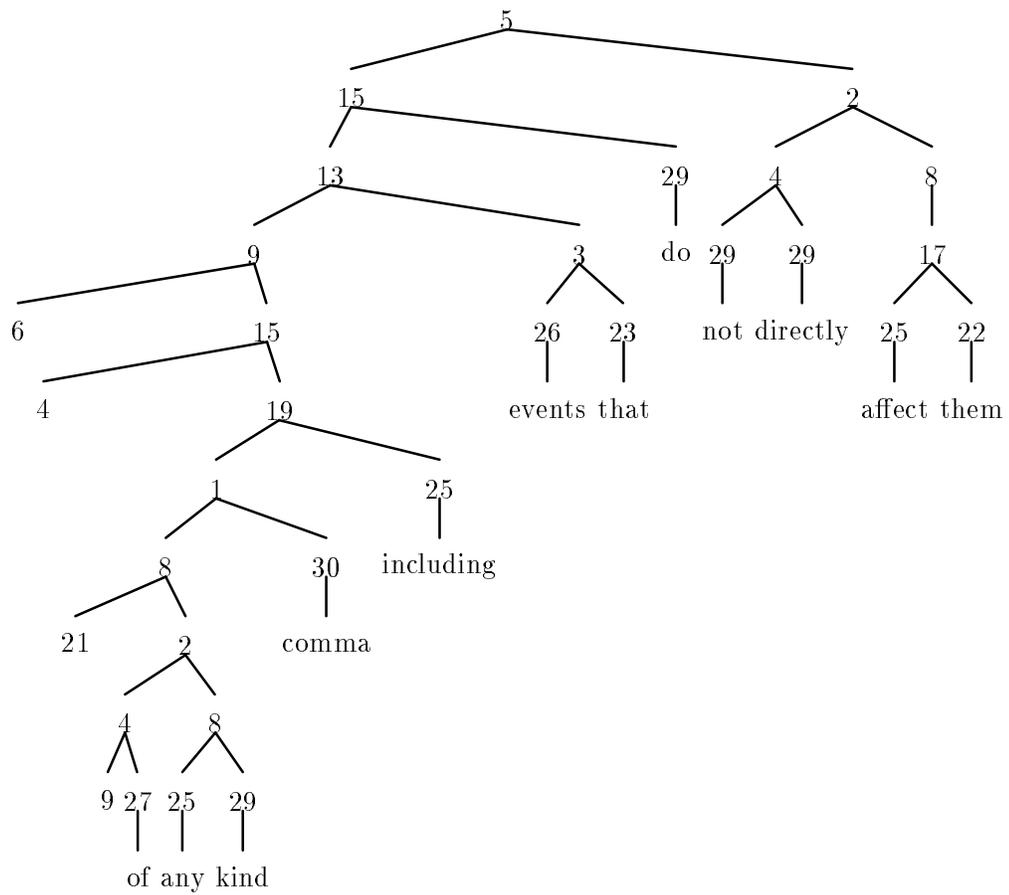

Figure 5: A parse tree for the part of a sentence shown in Figure 4 after another one hundred generations have passed.

These parse trees suggest that LUST fails to produce in one or two hundred generations the quality of parses that would make us leave human-generated parses to use machine-generated parses. It is likely that running LUST for thousands of generations will produce grammars that are closer to human-developed grammars or that may begin to approach an optimal grammar. Producing an optimal grammar will require a greater amount of computing power than is commonly available. Better results might also have been obtained if learning was based on a fully tagged set of terms, making additional (conventional) assumptions about parts-of-speech.

# 8 Conclusions

The performance of systems filtering documents into two groups, documents of probable interest and those the user probably does not want to see, may be improved if document components are labeled as to their grammatical parts-of-speech. This will make more precise the relationships between terms, allowing statements like "boy bites dog" and "dog bites boy" to be differentiated when searching for "boy" as biter, "boy" as bitten, and so forth. While scholars have studied disambiguation using grammatical techniques consistent with traditional linguistic parts-of-speech and syntactic rules (Burgin & Dillon, 1992), the use of genetic algorithms such as the one provided by LUST produce grammars optimized for the particular filtering and retrieval application of interest, as well as for a particular sublanguage (Bonzi, 1990; Damerau, 1990; Grishman & Kittredge, 1986; Haas & He, 1993; Losee & Haas, 1995). Learning the characteristics of a sublanguage has an obvious utility in supporting the discrimination between documents from different disciplines or with different stylistic characteristics (e.g. academic research vs. general non-fiction, or literature reviews vs. more traditional research articles).

The data presented here shows that the LUST system learns grammatical rules and part-of-speech tags, improving the quality of the initial randomly generated syntactic rules and part-of-speech labels. An original assumption of the author, that filtering performance could be used as a fitness function, measuring the quality of the parsing produced by a grammar, has proved to have little supported. As with the earlier work of Burgin and Dillon (1992), relatively little improvement was noted when additional linguistic knowledge becomes available about document components, although any filtering or retrieval improvement is welcome, no matter how small! While using retrieval performance as a fitness function does work, we believe that other functions will allow for more rapid learning of the characteristics of natural language. AMPL appears to be very useful in producing the desired knowledge, allowing us to produce genes that result in improved retrieval and allow us to better capture the rules of a grammar.

Experimental results suggest that further research might want to address several considerations. Future genetic algorithm systems supporting information retrieval and filtering need to use multiprocessing systems or parallel processing arrays if they are to allow for the tens of thousands of generations necessary if both syntactic rules and parts-of-speech are to be studied and accurately learned. In addition, larger databases need to be used in experiments if a greater variety of linguistic structures are to be incorporated into the grammar. This will require more rapid parsing.

If one is willing to assume that parts-of-speech are known accurately, the learning of syntactic rules can occur at a much higher rate than that experienced here. These tags might be provided by existing taggers. A compromise would be to accept the parts-of-speech for many but not for all terms. For example, following the technique used above, most or all terms might be forced into certain human developed categories initially, allowing the system to change the categories if it was "fit" to do so. Certain terms might be categorized as being of the same part-of-speech, without placing any semantic restrictions on the nature of this part-of-speech. For example, *run* and *jump* have a certain similarity; suggesting to the system that this similarity holds may prove a powerful form of prior knowledge without imposing additional grammatical ideas of what a "verb" is on the system.

Formal analytic techniques have been developed relating queries and filter characteristics to retrieval performance (Losee, 1995). This model needs to be expanded to include the formal characteristics of a grammar and the effects of grammatical tagging and disambiguation on performance. A stochastic model of the production of the grammar through evolutionary procedures may assist us in developing a formal model of all aspects of the LUST system.